\documentclass[letterpaper, 10 pt, conference]{ieeeconf}  

\IEEEoverridecommandlockouts                              

\usepackage{times} 
\usepackage{amsmath} 
\usepackage{amssymb}  

\usepackage{xcolor} 

\usepackage{booktabs, multirow} 
\usepackage{float}
\let\labelindent\relax
\usepackage{enumitem}
\usepackage{graphicx}
\usepackage{bbm}
\usepackage{dsfont}
\usepackage{wrapfig}
\usepackage{pifont}
\usepackage{xcolor}

\usepackage{enumitem}

\usepackage{subcaption}

\newcommand{\cmark}{\ding{51}}%
\newcommand{\xmark}{\ding{55}}%

\title{\LARGE \bf
FalconWing: An Ultra-Light Indoor Fixed-Wing \\ UAV Platform for Vision-Based Autonomy
}

\author{Yan Miao$^{1}$, Will Shen$^{1}$, Hang Cui$^{1}$ and  Sayan Mitra$^{1}$
\thanks{$^{1}$University of Illinois at Urbana-Champaign}
}

\begin{document}

\maketitle
\thispagestyle{empty}
\pagestyle{empty}

\begin{abstract}
We introduce \emph{FalconWing}, an ultra-light (150 g) indoor fixed-wing UAV platform for vision-based autonomy. 
Controlled indoor environment enables year-round repeatable UAV experiment but imposes strict weight and maneuverability limits on the UAV, motivating our ultra-light FalconWing design. 
FalconWing couples a lightweight hardware stack (137\,g airframe with a 9\,g camera) and offboard computation with a software stack featuring a photorealistic 3D Gaussian Splat (GSplat) simulator for developing and evaluating vision-based controllers.
We validate FalconWing on two challenging vision-based aerial case studies.
In the leader-follower case study, our best vision-based controller, trained via imitation learning on GSplat-rendered data augmented with domain randomization, achieves 100 \% tracking success across 3 types of leader maneuvers over 30 trials and shows robustness to leader's appearance shifts in simulation. 
In the autonomous landing case study, our vision-based controller trained purely in simulation transfers \emph{zero-shot} to real hardware, achieving an 80\% success rate over ten landing trials. 
%
We will release hardware designs, GSplat scenes, and dynamics models upon publication to make FalconWing an open-source \emph{flight kit} for engineering students and research labs.
\end{abstract}


\section{Introduction}

Autonomous fixed-wing UAVs are useful in applications such as for delivery~\cite{8701196}, navigation~\cite{wueest2018accurate, doi:10.2514/1.G006645}, and environmental monitoring~\cite{boon2017comparison} due to their energy efficiency and long endurance. 
Vision-based control \cite{miao2025zeroshotsimtorealvisualquadrotor, low2024sousvide} is important in GPS-denied zones, but it is challenging for fixed-wing UAV: it must maintain airspeed to generate lift; it is governed by nonlinear aerodynamics; and the onboard video stream could degrade due to vibration and turbulence.

Existing work addresses these challenges using relatively large \cite{aerospace8080228, s22176549, 8243120} and sensor-rich platforms \cite{wueest2018accurate, doi:10.2514/1.G006645} with GPS / GNSS, lidar, high-resolution cameras and onboard computation. 
While these platforms are suitable for large-scale outdoor experiments, such experiments typically require regulated airspaces and are constrained by weather and time-of-day limitations, reducing accessibility and experimental throughput. 
In contrast, indoor spaces, such as our 40$\times$20$\times$5m flying arena (Figure \ref{fig:gsplat}) and typical university gyms, can provide weather/time-independent environment, which enables more frequent and accessible experiments under controlled perturbations (e.g., fan-generated wind).

\begin{figure}[htbp]
  \centering
  \includegraphics[width=\linewidth]{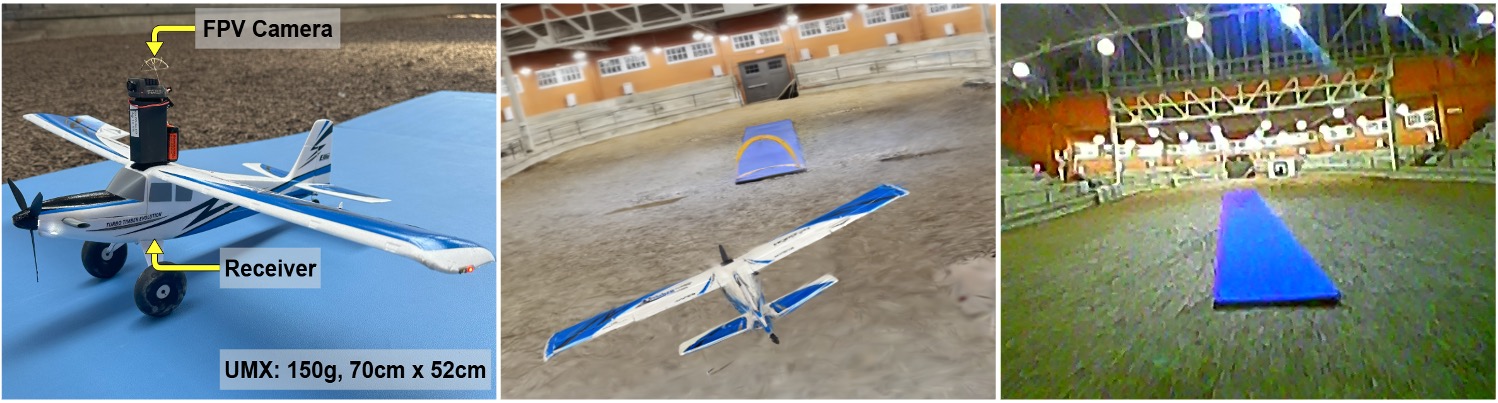}
  \caption{\small{\textbf{Left:} Our ultra-light 150\,g fixed-wing aircraft for indoor aerial research, equipped with a FPV camera and ROS-enabled autonomous control. \textbf{Middle:} Onboard view for leader-follower visual tracking using a digital camera. \textbf{Right:} Onboard view during autonomous landing using an analog camera.}}
  \label{fig:UMX}
\end{figure}

\begin{figure*}[htbp]
    \centering
    \includegraphics[width=0.85\textwidth]{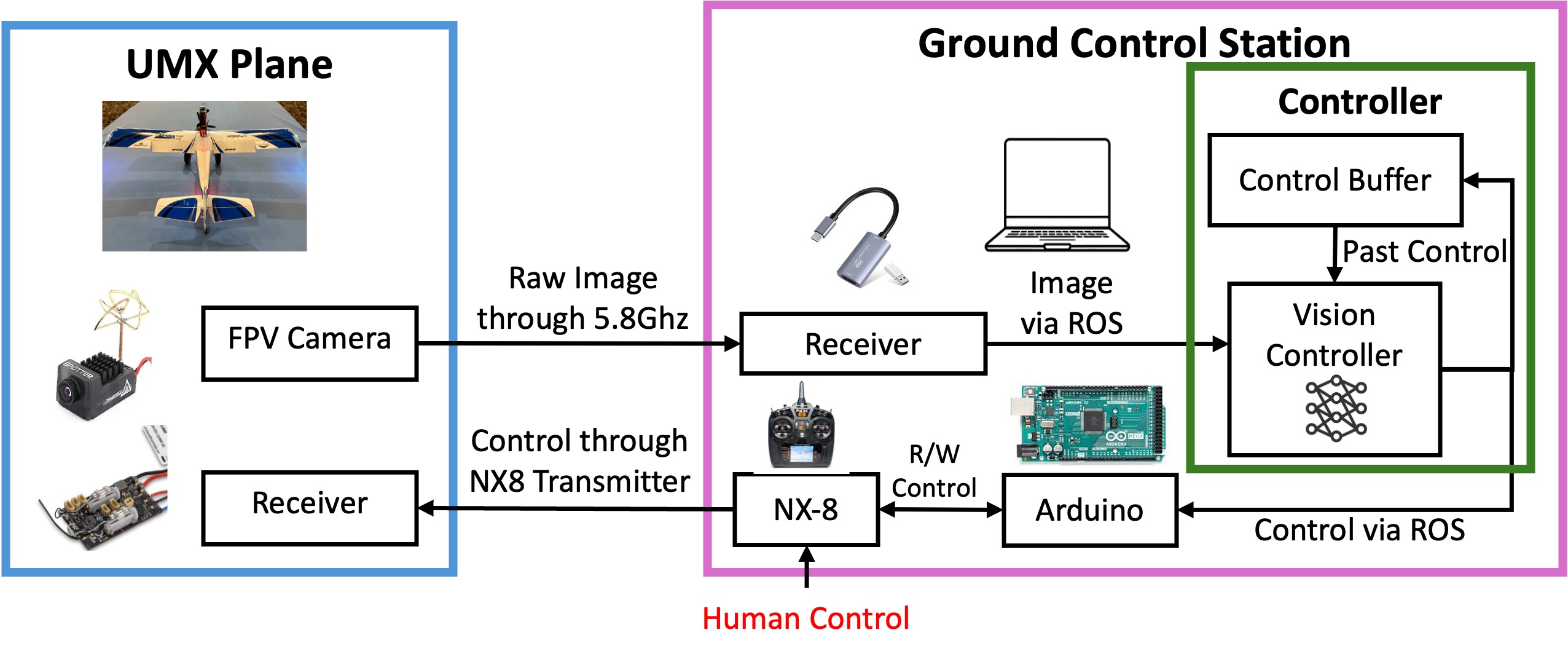}
    \caption{\small{Architecture of FalconWing Hardware: a light 9\,g FPV camera mounted on the fixed-wing plane streams images to the ground control station, where images are published to ROS. The controller reads published image plus buffered past controls, computes new flight control, and sends it via ROS to an Arduino. The Arduino writes these commands into the Spektrum NX-8 trainer port, closing the vision-based control loop over radio. The human pilot can instantly reclaim control at any time via a transmitter switch.}}
    \label{fig:system}
\end{figure*}

Indoor flight, however, imposes tight weight and maneuverability constraints: every additional gram raises the minimum required airspeed and thus increasing the minimum turning radius. 
A 150g aircraft requires approximately 7m/s minimum airspeed and an 8.7m minimum turning radius (as equations shown in Appendix); while a 300g aircraft (e.g. adding a 174g Jetson Orin) would require 10m/s minimum airspeed and 17.5m turning radius. 
This shows that existing heavy and sensor-rich fixed-wing UAV platform \cite{wueest2018accurate} with onboard computation ($\sim$890g) are often impractical to maneuver within a small space (e.g. 20\,m indoor width) and leaves an important yet under-explored design gap for a lightweight sensor-minimal fixed-wing UAV suitable for iterative and reproducible indoor experiments.

To address this gap, we introduce \emph{FalconWing}, a vision-based fixed-wing aircraft research platform weighing just 150\,g with a 9\,g FPV camera and offboard computation. FalconWing couples a sensor-minimal hardware stack supporting both manual and autonomous modes (Section~\ref{sec:hardware_system}) with a software suite comprising a photorealistic Gaussian Splat (GSplat) simulation environment (Section~\ref{sec:gsplat}) and a system-identified nonlinear dynamics model (Section~\ref{sec:sysid}) used for training.
We use GSplat \cite{kerbl3Dgaussians} as the simulation framework due to its photorealistic environment and its past success in sim-to-real deployment of the vision-based controller in quadrotor navigation \cite{miao2025zeroshotsimtorealvisualquadrotor, low2024sousvide}. 
%

To demonstrate FalconWing's capability, we tackle two challenging aerial case studies using only onboard vision: leader-follower visual tracking in simulation (Section~\ref{sec:leader-follower}) and zero-shot sim-to-real transfer of a vision-based autonomous landing controller in indoor environments (Section~\ref{sec:autonomous_landing}).
In the visual-tracking case study, we train a vision-based controller through imitation learning on a state-based expert controller in simulation and remove reliance on known target states and IMU, required by prior work \cite{miao2025zeroshotsimtorealvisualquadrotor}.
We also apply GSplat-level domain randomization to the training dataset to improve robustness.
Experiments in simulation show that our vision-based controller can follow the leader under 3 types of unseen maneuvers with 100\% success over 30 trials and shows certain robustness to leader appearance and size changes.
In the autonomous-landing case study, experiments show that the vision-based controller trained purely in simulation using the same approach as the visual tracking can transfer \emph{zero-shot} to hardware, achieving an $80\%$ success over ten trials.

Although autonomous vision-based control of UAV could be straightforward with rich multi-sensor stacks or motion-capture infrastructure, achieving reliable vision-based control with an ultra-light sensor-minimal fixed-wing platform indoors is challenging due to tighter workspace and limited sensing.
In summary, our contributions are:
(i) \emph{FalconWing platform:} an ultra-light indoor fixed-wing UAV platform paired with a GSplat simulation and system-identified dynamics. We envision FalconWing as an accessible ``flight kit'' for undergraduate engineering courses and research labs, where students can gain hands-on experience with airframe assembly, ROS-based vision pipeline setup, and Arduino-based MCU programming;
(ii) \emph{Demonstration on two challenging case studies:}
FalconWing platform is capable of developing and testing vision-based controllers for challenging tasks: leader-follower visual tracking and autonomous landing. 
 
\section{Related Work}

\paragraph{Aerial Autonomy}  
Traditional approaches to aerial autonomy rely heavily on external sensors such as GPS and IMUs for state estimation and control. 
For example, \cite{wueest2018accurate} demonstrated accurate vision-aided navigation using GNSS, IMU-assisted Kalman filtering, while \cite{doi:10.2514/1.G006645} achieved agile maneuvering with high-precision motion capture systems. 
Some work tries to mitigate reliance on tracking sensors by adding learning-based perception modules like faster R-CNN \cite{8243120} for pose estimation but still requires high-fidelity digital cameras. 
\cite{10.1007/s10846-009-9382-2, barber2009vision, 6476736} have attempted vision-based flight for landing, 
but all of those platforms rely on large, heavy platforms (e.g., 1-5kg) with multi-sensor suites, making them impractical for indoor or GPS-denied scenarios.


\paragraph{Photorealistic Scene Representation in Robotics}
Photorealistic scene representations like Gaussian Splat \cite{kerbl3Dgaussians} have recently emerged as powerful tools for 3D reconstruction. 
Variants and extensions have been applied to diverse robotics tasks, including 3D scene editing~\cite{ni2025efficientinteractive3dmultiobject}, pose estimation~\cite{yen2020inerf, bortolon20246dgs} and navigation~\cite{nerf-nav}.
RialTo~\cite{torne2024rialto} demonstrates the potential real-to-sim-to-real transfer for robot arm manipulation, by constructing a simulation based on real images, and deploy simulation-trained model to real world again. 
\cite{miao2025zeroshotsimtorealvisualquadrotor,low2024sousvide} achieves zero-shot drone navigation using policies trained in photorealistic scene representation.

%


\section{FalconWing Hardware Stack}
\label{sec:hardware_system}

In this section, we introduce FalconWing's 150g ultra-light hardware stack (Figure \ref{fig:system}) designed for indoor flights.

\paragraph{Base Airframe}  
The platform builds on the UMX Turbo Timber\textsuperscript{\textregistered} (Figure~\ref{fig:UMX} Left), a hobby-grade airframe chosen for its lightweight design (137\,g), integrated electronic speed controller (ESC) and flight controller, and off-the-shelf parts availability. 
Its durable foam fuselage withstands crashes during iterative testing, while the 70\,cm wingspan with flaps balances maneuverability and provides extra lift. 

\paragraph{Vision System}  
A 9\,g RunCam Spotter\textsuperscript{\textregistered} analog FPV camera provides onboard vision. 
Mounted along the fuselage centerline via a custom 3D-printed bracket (5\,g), the camera avoids propeller occlusion and preserves the center of gravity. 
Images are transmitted via a 5.725\,GHz analog link to a ground control station, where a receiver forwards the signal to a USB capture card. 
The card streams 640$\times$480 RGB frames at 20\,Hz into a ROS topic, enabling real-time processing. 
An LC filter is also connected to the camera to reduce high-frequency noise from motor vibrations.
We also provide the option to switch to a digital camera if users value image quality over lightweight.

\paragraph{Control Interface}  
Autonomous flight is enabled through the Spektrum NX-8\textsuperscript{\textregistered} transmitter’s trainer port (Figure~\ref{fig:system}). 
An Arduino Mega 2560 bridges ROS and the transmitter via rosserial-python, translating four-channel pulse-position modulation (PPM) signals (throttle, aileron, elevator, and rudder) between ROS messages (20\,Hz) and the transmitter’s serial interface.
To minimize aircraft weight for indoor flights, all computation runs offboard on a ground station with an RTX 4090 GPU.

\paragraph{Operating Modes}  
We configure two modes:  
\begin{itemize}[leftmargin=*]
    \item \textbf{Manual Mode:} A human pilot manually flies via the NX-8. The Arduino logs pilot commands and time-synchronized images to ROS to build datasets for later system identification and controller training. Since we retain the 13.5\,g Horizon Hobby receiver (Figure \ref{fig:system}) and its integrated flight controller with IMU, expert pilots can perform manual flight and aerobatics. Switching to Autonomous Mode requires only a single transmitter toggle.
    \item \textbf{Autonomous Mode:} The Arduino subscribes to the ROS topic publishing autonomous control commands from the vision-based controller and writes these commands to the trainer port, closing the vision-based control loop. The human pilot can instantly take over by flipping the same transmitter switch whenever intervention is required.
\end{itemize}
These modifications transform a hobbyist airframe into a ROS-compatible platform for indoor vision-based fixed-wing research: Manual Mode is used to generate training data, and the same hardware supports closed-loop Autonomous Mode experiments via a single switch.

\paragraph{Safety Mechanisms}  
To mitigate the risk of degraded analog video during Autonomous Mode, we deploy a frame-quality monitor based on the Structural Similarity Index (SSIM)~\cite{1284395}. More specifically, if the SSIM between consecutive frames falls below an empirical threshold (0.7 for five consecutive frames, values set by empirical testing), we raise a flag alerting human pilots to take control immediately, since that usually indicates severe analog noise.

\paragraph{Open-source Availability \& Educational Purposes}
A key advantage of our lightweight, sensor-minimal design is fast assembly and maintenance. 
With our part list and step-by-step 15-page manual (to be released upon publication), new users can assemble and fly FalconWing in under 10 hours. Minor to medium crash repairs are inexpensive and quick (e.g., ESC \$80, propeller \$8; swap time $\approx$ 2 hours), minimizing downtime.
In addition, FalconWing could be suited for high-school and undergraduate courses as educational “flight kits”, where students can gain hands-on experience with both hardware assembly and software development.
Just as F1-TENTH \cite{pmlr-v123-o-kelly20a} has inspired ground-robotics education, we envision FalconWing becoming the go-to aerial platform for teaching and research.

\section{FalconWing Software Stack}
\label{sec:software_systm}

In this section, we introduce FalconWing's software stack, which includes two variations of photorealistic simulation environment (Section \ref{sec:gsplat}), identified airplane dynamics used for simulation training (Section \ref{sec:sysid}) and the open-source availability (Section \ref{sec:open-source-code}).

\subsection{Photorealistic Simulation via Gaussian Splatting}
\label{sec:gsplat}

\begin{figure}[htbp]
    \centering
    \includegraphics[width=\linewidth]{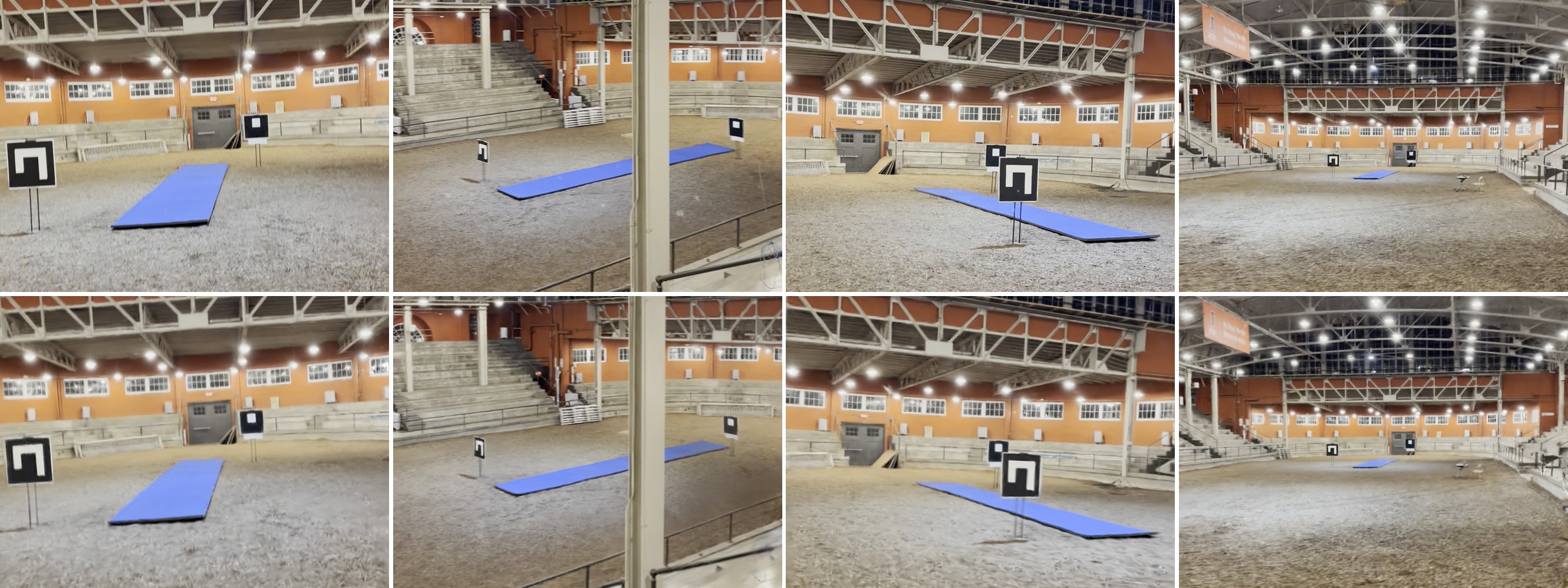}
    \caption{\small{FalconWing's simulation can render photorealistic images using Gaussian Splat from different poses. The top row shows 4 real world images in our flying arena, while the bottom row displays corresponding images rendered by GSplat at the same coordinates.}}
    \label{fig:gsplat}
\end{figure}

A photorealistic simulation environment can help mitigate the sim-to-real gap in vision-based control. 
In FalconWing, we synthesize a photorealistic simulation environment $G$, that can render a photorealistic image $I$ from any virtual camera pose $p$ in world coordinates, i.e., $I = G(p)$. 
We achieve this by enhancing FalconGym \cite{miao2025zeroshotsimtorealvisualquadrotor}, replacing Neural Radiance Fields (NeRF)\cite{10.1145/3503250} with Gaussian Splat (GSplat) \cite{kerbl3Dgaussians} for faster and better rendering and eliminating the need for a motion capture system. 

\paragraph{Data Collection and Calibration}
We used the onboard FPV camera to capture about 2000 images throughout our $40\times20\times5$m indoor arena from different positions. 
Camera intrinsics and poses are estimated using COLMAP~\cite{7780814}. 
We calibrate camera poses to the world frame by placing an 80 cm ArUco marker beside the runway (Figure \ref{fig:gsplat}). 
Using OpenCV’s ArUco detector, we identify the marker center in a subset of images and treat it as the global origin. 
We then compute the rigid transform between COLMAP and world frames via the Kabsch-Umeyama algorithm~\cite{Kabsch:a15629}.

\paragraph{GSplat-based Simulation Construction}
With calibrated poses, we feed the images and transforms into the open-source NeRFStudio Splatfacto pipeline~\cite{nerfstudio}. 
On an NVIDIA RTX 4090, training converges in approximately 15 minutes. 
The resulting model supports fast rendering with an average of 0.004s for a 960x720 image.

\paragraph{Digital Camera Variant Simulation}
To accommodate researchers who favor image quality over minimal mass, we repeat the above procedure using a digital ArduCam RGB camera. 
This provides an additional digital camera-based simulation environment. 
Figure~\ref{fig:gsplat} qualitatively compares real-world images with renders from simulation environment.

\paragraph{UMX Gaussian Splatting}
\label{sec:editable-gsplat}
Utilizing the same techniques, we also construct our UMX plane as a GSplat asset in the simulation. 
With the Edit API in \cite{miao2025performanceguidedrefinementvisualaerial}, we can place and render a photorealistic leader aircraft in simulation for the leader-follower visual tracking case study (Section~\ref{sec:leader-follower}). 

\subsection{Nonlinear System Identification}
\label{sec:sysid}

With the photorealistic simulation established in Section~\ref{sec:gsplat}, we now aim to obtain a reliable dynamics model of our FalconWing aircraft. 
We adopt a reduced-order kinematic model inspired by standard fixed-wing dynamics formulations~\cite{doi:https://doi.org/10.1002/9781119174882.ch1}. 
Specifically, we represent the aircraft state as \( x = [p_x, p_y, p_z, \theta, \gamma, \phi, v_x, v_y, v_z] \), which captures the aircraft's position, orientation (pitch, yaw, roll), and linear velocity. 
Note the first 6 state variable is exactly camera (plane) pose $p$. Our control inputs are defined as \( u = [u_T, \delta_a, \theta_c, \gamma_c] \), corresponding to throttle, commanded aileron, elevator and rudder. 
We model the discrete-time nonlinear dynamics as a parametric function \( f_K \), i.e., $f_K(x_t, u_t)$ where \( K \) denotes the vector of unknown dynamics parameters we seek to estimate. 

\begin{figure}[htbp]
    \centering
    \includegraphics[width=\linewidth]{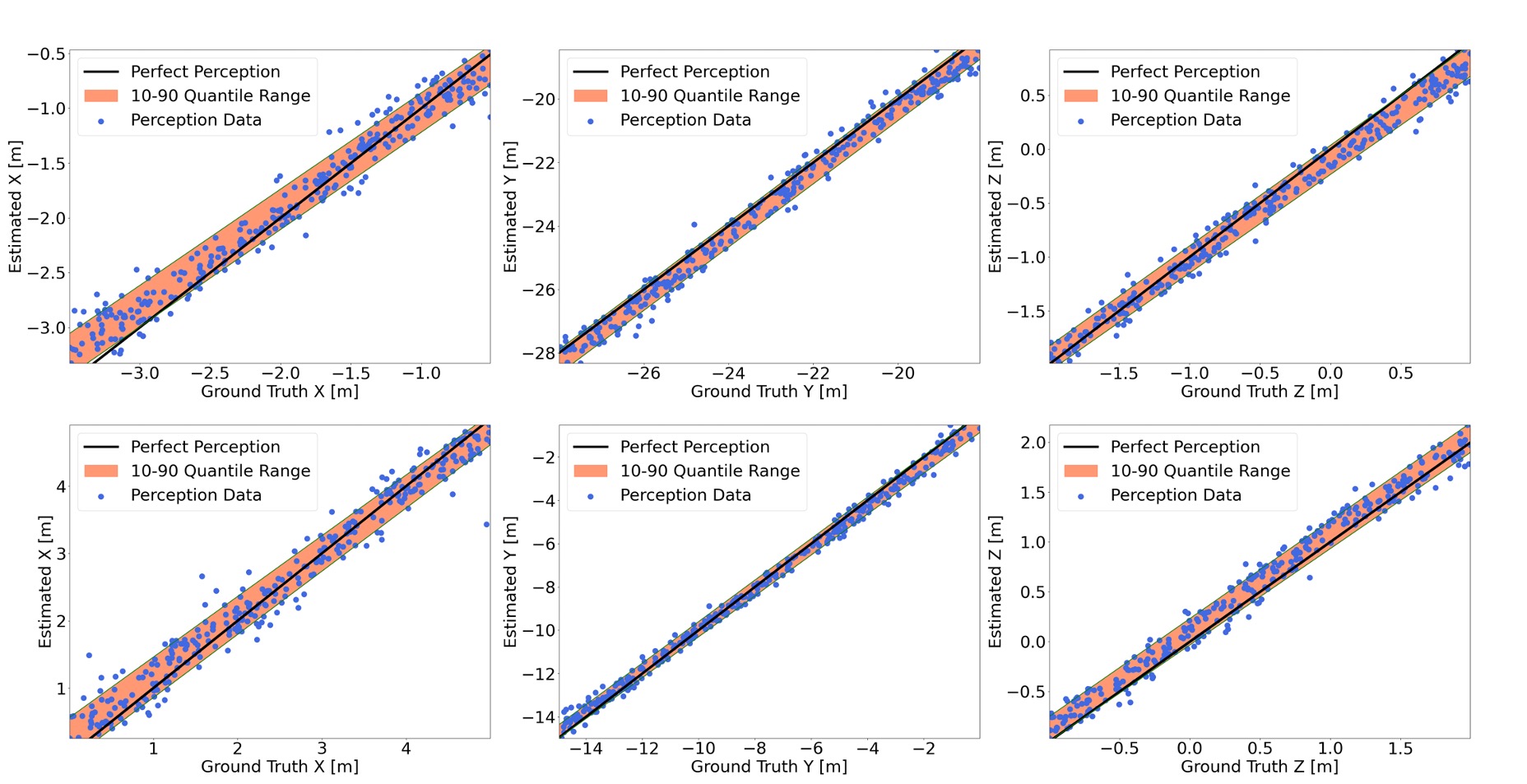}
    \caption{iGSplat performance: top row shows one-shot pose estimation in analog-camera based simulation, while bottom row shows result in the digital-camera based simulation.}
    \label{fig:igsplat}
\end{figure}

\paragraph{Hybrid State Estimation}  
Since no motion capture system is yet available in our flying arena, we must estimate ground-truth states purely from vision input. 
We propose a hybrid vision-based state-estimation pipeline: when the aircraft is close enough to the ArUco marker and it is detectable by the OpenCV ArUco library, we directly use its estimates; otherwise, inspired by the NPE model in FalconGym \cite{miao2025zeroshotsimtorealvisualquadrotor}, we utilize a neural network-based inverse Gaussian Splat (iGSplat) model to infer camera poses from single RGB frames. 
Although iterative pose-optimization methods such as iNeRF~\cite{yen2020inerf} can estimate camera poses accurately, they are computationally expensive. 
Therefore, we train a single-shot neural network architecture for efficient inference.
Specifically, our iGSplat model employs a Vision Transformer (ViT) backbone pretrained on ImageNet-21k~\cite{ridnik2021imagenetk}. 
We freeze early transformer layers to leverage general visual feature extraction, adding a trainable regression head for direct camera-pose estimation. 
To circumvent discontinuities inherent in angular regression, our network predicts sine and cosine values of pitch, yaw, and roll angles, subsequently recovering angular orientations via a trigonometric transformation. The qualitative result of iGSplat on 200 unseen images can be found in Figure \ref{fig:igsplat}, which shows a relatively accurate pose estimation using both an analog camera and a digital camera, with an average of 0.42m position estimation error and 2.37 degrees yaw estimation error.

\paragraph{System Parameter Identification through Least Square}
We collect a dataset $\mathcal{D}_I=\{(I_t,u_t)\}$ by recording images and pilot inputs during Manual Mode (Section~\ref{sec:hardware_system}). 
Applying our hybrid estimator yields pose sequences $\{p_t\}$, which we differentiate to obtain velocity and form state-action pairs $\mathcal{D}_x=\{(x_t,u_t)\}$, where frames corrupted by significant analog noise or yielding clearly implausible pose estimations are manually removed. 
Then we solve
\[
K^* = \arg\min_K \sum_{(x_t, u_t)\in \mathcal{D}_x}\bigl\|f_K(x_t,u_t)-x_{t+1}\bigr\|_2^2
\]
via nonlinear least square (SciPy). The optimized parameter set $K^*$ yields a reliable dynamics model for controller design and training in simulation.

\subsection{Open-source Software Package}
\label{sec:open-source-code}

In addition to the hardware part list and user manual, we also plan to open-source the complete FalconWing software stack including: two photorealistic simulation environments (analog and digital camera variants) and system-identified dynamics parameters with everything packed in a conda environment for easy distribution. 
This digital twin is designed as an open-source reusable benchmark for future research in vision-based fixed-wing control.

We next demonstrate FalconWing's capabilities through two challenging aerial case studies: leader-follower visual tracking (Section \ref{sec:leader-follower}) and vision-based autonomous landing (Section \ref{sec:autonomous_landing}). 
Note that although our FalconWing hardware (Section \ref{sec:hardware_system}) does carry a self‐leveling flight controller, which researchers may choose to employ in their applications, we deliberately disable the autopilot assists for both case studies so as to isolate pure vision‐based controller performance. 
\section{Case Study: Leader-Follower Visual Tracking}
\label{sec:leader-follower} 

\begin{figure}[htbp]
  \centering
  \includegraphics[width=\linewidth]{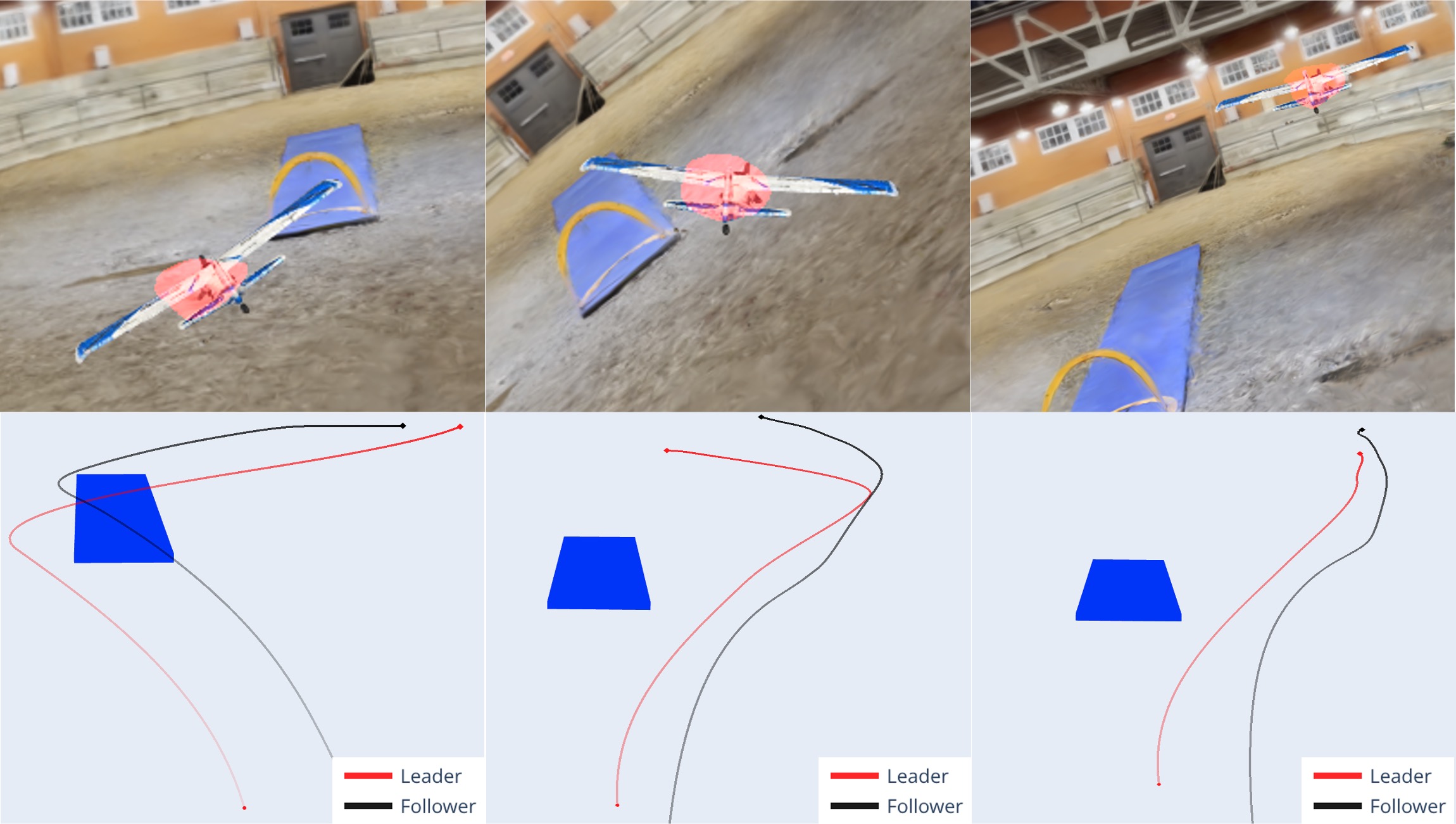}
  \caption{\small{Onboard camera views and trajectory plots for the leader-follower case study: our vision-based controller on the follower can closely track the leaders in three different leader maneuvers. The annotated red part on the onboard images indicating the mask detection result described in Section \ref{sec:UMX-detection}.}}
  \label{fig:leader_chaser_traj}
\end{figure}

\begin{figure*}[htbp]
  \centering
  \includegraphics[width=0.75\linewidth]{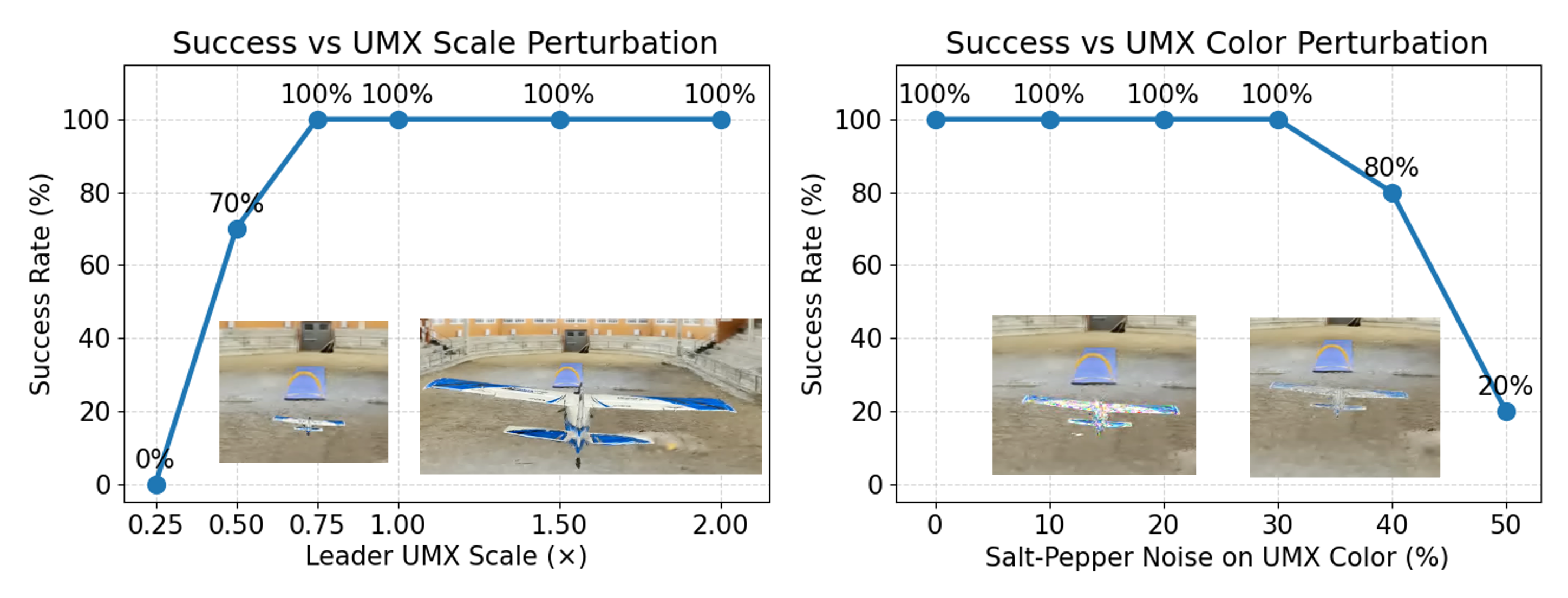}
  \caption{\small{Our "RGB+Mask" vision-based controller is relatively robust to both leader scale and color perturbation in the GSplat. For each perturbation level, we ran 10 experiments with slightly different initial conditions and report the average success rate.}}
  \label{fig:ablation}
\end{figure*}

In this case study, we consider the problem of visual tracking, where the follower UMX aircraft needs to track a leading UMX aircraft using vision.
%
Fixed-wing leader-follower visual tracking is vital for tasks such as search‐and‐rescue, delivery and aerial navigation.
%
Yet visual tracking is challenging due to the small size of the aircraft (in our case, 70cm$\times$52cm) in the image space, its nonlinear underactuated dynamics, and the lack of ground‐truth state feedback. 
\cite{li2024visualtrackingintermittentvisibility} tackles a similar tracking problem, but they attach an ArUco marker to the leader aircraft that significantly reduces perception difficulty, while we design controllers that directly tracks the leader aircraft via RGB images.

In the following subsections, we develop three distinct neural visual tracking controllers and evaluate their ability to track under three types of leader representative maneuvers: a left‐turn S‐shape descent, a right‐turn S‐shape ascent, and a right‐turn sharp climb, as shown in Figure \ref{fig:leader_chaser_traj}. 
As shown in Table \ref{tab:leader-follower-analysis}, we measure the tracking performance using 3 key metrics: (i). Success Rate (SR), fraction of trials where the follower maintains visual lock on the leader throughout all frames; (ii) Average Tracking Error (ATE), defined as the average displacement between the leader and the follower minus the initial tracking offset; (iii) Average Runtime (ART), per‐frame inference time of the controller using the 4090 GPU of the ground control station.

\subsection{Vision Controller: Direct Imitation Learning}
\label{sec:controller-training}
Building on FalconGym’s ~\cite{miao2025zeroshotsimtorealvisualquadrotor} success in quadrotor navigation via imitation learning, we started by designing an end-to-end vision-based controller that maps onboard RGB images directly to fixed-wing control commands, while addressing three key limitations of FalconGym: (i) reliance on known target positions, (ii) dependence on IMU readings, (iii) heavyweight dual‐ViT architecture.
To overcome these constraints, our single ViT-based network ingests only the current image $I_t$ and a history of the past 30 control inputs $u_{t-30:t-1}$, which implicitly encode temporal state information and fuses them via a lightweight self-attention module (green box, Figure~\ref{fig:system}), thereby eliminating explicit pose estimation and IMU usage while reducing model size for faster inference time. 
We try out different length of history and find 30 being a suitable length of that balances model size and effectiveness.

We train this multi-modal controller $\pi_{\psi}(I_t, u_{t-30:t-1})$ by learning from an expert. The expert state-based controller $\pi^*$ is first implemented following standard fixed-wing designs~\cite{doi:https://doi.org/10.1002/9781119174882.ch4}. 
Then we configure the leader UMX plane to always use the expert state-based controller to follow predefined waypoints. 
During training data collection, we also configure the follower plane to use the expert state-based controller that has access to both leader's ground truth state $\hat{x}_t$ and its own ground truth state $x_t$ to compute optimal actions $ u^*_t = \pi^*(x_t, \hat{x}_t) $. 
To broaden the state-action distribution and improve robustness, we inject mild Gaussian noise, $ u_t = u_t^* + \mathcal{N}(0,\sigma^2), $ into the follower's expert outputs, encouraging slight exploration of off-nominal trajectories without compromising feasibility. 
We render the image $I_t$ with the leader UMX plane at each state $x_t$ using the UMX asset (Section \ref{sec:editable-gsplat}) in FalconWing's photorealistic simulation and log the expert history to form the dataset $ \mathcal{D}_C = \bigl\{(I_t,\,u_{t-30:t-1},\,u_t^*)\bigr\}. $ 
We collect this dataset by setting random but dynamically feasible waypoints for leader UMX to explore. 
We then optimize the controller parameters $\psi$ by minimizing the mean squared error between predicted and expert actions:
\[
\mathcal{L}(\psi)
= \frac{1}{|\mathcal{D}_C|}\sum_{(I_t,u_{t-30:t-1},u_t^*)\in\mathcal{D}_C}\bigl\|\pi_{\psi}(I_t,u_{t-30:t-1}) - u_t^*\bigr\|_2^2.
\]

We refer to this controller as ``RGB'' in Table~\ref{tab:leader-follower-analysis}, because the visual input to the controller is an RGB image. While ``RGB'' attains high SR and low ATE on the training trajectories, it fails to generalize to unseen leader maneuvers. 
This is consistent with observation in FalconGym~\cite{miao2025zeroshotsimtorealvisualquadrotor} that a controller trained in one scenario tends to specialize in that single scenario. 
We believe this overfitting failure mode happens because the leader UMX occupies only a small fraction of image pixels, so the network overfits to background appearance rather than learning a transferable representation of the leader aircraft.

\subsection{Vision Controller: UMX Mask Detection}
\label{sec:UMX-detection}

To improve generalization and avoid overfitting, we decouple perception from control: a UNet \cite{10.1007/978-3-319-24574-4_28} first predicts a binary mask of the leader from the onboard RGB image; a lightweight ResNet then consumes the mask together with the past control history to output the next action. We denote this approach as ``Mask'' in Table \ref{tab:leader-follower-analysis}.

\begin{figure}[htbp]
  \centering
  \includegraphics[width=0.65\linewidth]{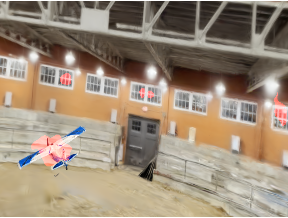}
  \caption{\small{An example of failed perception  that leads to downstream tracking error. The mask prediction confuses leading aircraft with the background window using ``Mask'' approach in Section \ref{sec:UMX-detection}.}}
  \label{fig:mask-failure}
\end{figure}

Training the UNet also leverages our UMX GSplat assets (Section~\ref{sec:editable-gsplat}). 
We synthesize the perception training dataset by sampling leader plane poses across the arena workspace and spawning the follower aircraft (camera) at feasible viewpoints (ensuring the leader is roughly front-facing). 
Because the poses of Gaussians corresponding to the leader’s wing tips and nose-tail endpoints are known and the camera matrix is calibrated, we can approximate the aircraft as a 3D ellipsoid and project it onto the image plane to obtain ground-truth masks using standard geometry with traditional computer vision techniques. 
We collect 10K pairs of RGB images and masked images to train for this UNet. 
The trained detection result is shown in Figure \ref{fig:leader_chaser_traj} (we annotate the mask as red on top of the plane for visualization purposes). 
After training the UNet, we apply the same imitation-learning setup as in Section~\ref{sec:controller-training} but train the controller to map mask and past controls to the current action. 
The training dataset consists of around 100K mask-action pairs (1000 trajectories). 
This ``Mask'' approach reduces reliance on background cues and yields a more generalizable leader-follower controller than ``RGB'', as shown in Table~\ref{tab:leader-follower-analysis}. 
However, it is not completely immune to perception errors: in one unseen maneuver, the UNet confuses the leading UMX with a bright window pattern (similar white stripes), producing an incorrect mask and steering the follower toward the background window, as shown in Figure \ref{fig:mask-failure}.

\subsection{Vision Controller: RGB + Mask}
To further mitigate such perception failures that cause downstream control problems, we introduce ``RGB+Mask,'' which stacks the predicted binary mask as a fourth channel on top of the RGB image and repeats the imitation-learning procedure. 
The additional appearance context helps both disambiguate false positives and reduces overfitting, improving both SR and ATE across both training and unseen maneuvers (Table~\ref{tab:leader-follower-analysis}). 
The trade-off is increased inference time due to two sequential networks (U-Net followed by a 4-channel ResNet), which makes hardware deployment at runtime risky.

\subsection{Domain Randomization}
\begin{figure}[htbp]
  \centering
  \includegraphics[width=\linewidth]{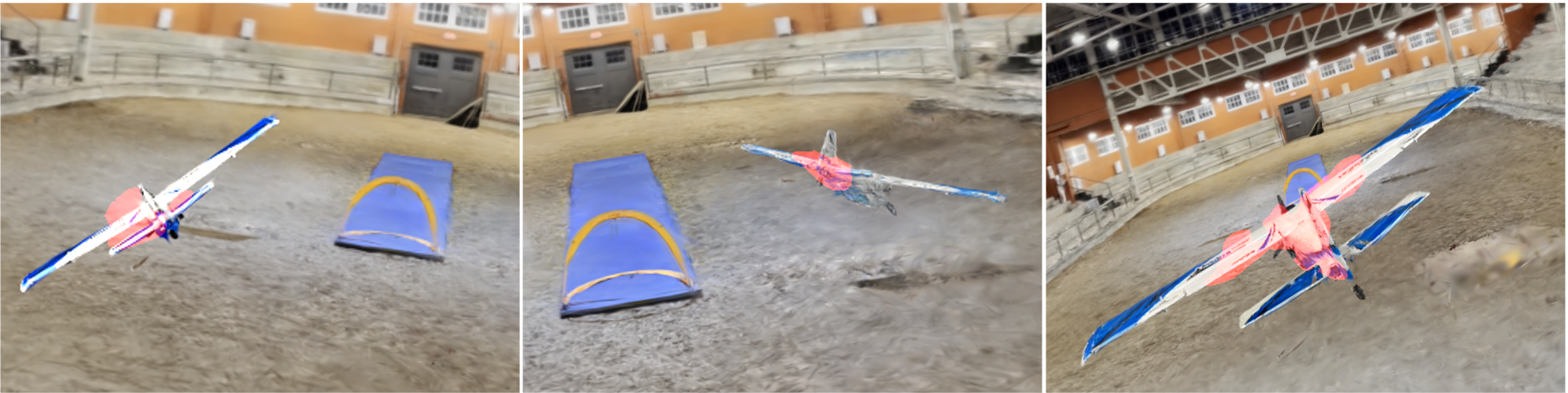}
  \caption{\small{We enable domain randomization in terms of leader gaussians' color and scale to improve detection robustness.}}
  \label{fig:domain-randomization}
\end{figure}

Beyond pose and camera sampling, we apply domain randomization to mask detection to enhance robustness and reduce sim-to-real gaps. 
Specifically, we perturb the leader UMX appearance by varying scales and colors of the gaussians associated with the leader’s UMX (Section \ref{sec:editable-gsplat}). 
Figure~\ref{fig:domain-randomization} illustrates the three perturbations (brightness, salt-pepper noise and scaling) used during training.

\subsection{Experiment Setup \& Tracking Performance Analysis}

Due to the lack of motion capture in the flying arena, we cannot safely run the leader plane with state-based control in real world. Therefore, for safety reasons, all leader-follower experiments are conducted in simulation.
Sim-to-real validation of FalconWing will appear in the next autonomous landing case study (Section~\ref{sec:autonomous_landing}).

\textit{Baseline.} We first evaluate a state-based follower that has access to ground-truth states of both planes and uses the same fixed-wing controller as the leader. For each of the three unseen leader maneuvers (a left‐turn S‐shape descent, a right‐turn S‐shape ascent, and a right‐turn sharp climb, shown in ~\ref{fig:leader_chaser_traj}), we run $10$ trials with slight variations in initial conditions and report the average SR, ATE, and ART over 30 trials in Table~\ref{tab:leader-follower-analysis}. As expected, the state-based baseline achieves the best SR and lowest ATE, and its ART is negligible because the computation is basically simple tensor operations.

\textit{Direct RGB Controller.} The ``RGB'' controller closely imitates the expert on its specific training trajectory but performs poorly to unseen leader maneuvers. 
Its key advantage is runtime: ART $\approx$ $0.02$\,s, which can easily fit our $20$\,Hz control loop and is therefore suitable for hardware deployment.

\textit{Mask-based Controller.} The ``Mask'' variant substantially reduces overfitting by conditioning control on the predicted UMX mask and past controls, improving SR and ATE across unseen maneuvers. However, it is susceptible to perception errors and has slightly higher runtime.

\textit{RGB+Mask Controller.} Stacking the predicted mask as a fourth channel (``RGB+Mask'') mitigates the rare mask failures while retaining the generalization benefits of the Mask approach. It delivers the strongest overall SR and ATE among learned policies, but at the cost of the highest ART, which complicates hardware deployment.

\begin{table}[htbp]\centering
\caption{\small{Controller Performance for Leader-Follower Case Study}}
\label{tab:leader-follower-analysis}
\scriptsize
\begin{tabular}{lcccc}\toprule
    Scenarios & Controller Input & SR\% $\uparrow$ & ATE [cm] $\downarrow$ & ART [s] $\downarrow$  \\\midrule
    Training & \underline{State-based} & \underline{100\%} & \underline{72}  & \underline{$\approx 0.00$} \\ 
    & RGB &  100\% & 78  & \textbf{0.02} \\
    & Mask & 100\% & 91 & 0.08 \\
    & \textbf{RGB+Mask} & \textbf{100\%} & \textbf{73} & 0.13\\ \midrule

    Unseen & \underline{State-based} & \underline{100\%} & \underline{79}  & \underline{$\approx 0.00$} \\
    & RGB & 30\% & 102 & \textbf{0.02} \\
    & Mask & 90\% & 139 &  0.08 \\
    & \textbf{RGB+Mask} & \textbf{100\%} & \textbf{94} & 0.13 \\ \bottomrule
\end{tabular}
\end{table}

\textit{Robustness.} We further probe robustness of our ``RGB+Mask'' approach using scale and salt-and-pepper perturbations applied at the Gaussian color space. As is shown in Figure \ref{fig:ablation}, across $10$ runs per perturbation condition, the controller remains reliable over a broad range of apparent sizes (0.5x to 2x): performance degrades only when the leader is reduced to $50\%$ of its nominal size, where detection becomes unreliable. The controller is also robust when salt-and-pepper noise injected onto 30\% of leader-associated Gaussians.




\section{Case Study: Vision-Based Autonomous Landing}
\label{sec:autonomous_landing}

In this section, we utilize the FalconWing platform  to tackle another challenging fixed-wing case study: vision-based landing and show success sim-to-real transfer. 
Landing is fundamental for all aerial applications and requires precise perception of the runway and tight control of glide slope. 
Even skilled RC (radio-controlled planes) pilots typically need weeks of practice to master consistent landings. 
We tackle vision-based landings to mimic landing in a GPS-denied zone where no ground-truth state is available.
\cite{s22176549} tackles a similar problem but assume landing zone has an obvious marker, while we work directly with the runway.

In the rest of this section, we describe our vision-based controller for landing (Section \ref{sec:landing-vision-policy}), indoor landing setup (Section~\ref{sec:experiment-setup}) and evaluate the vision-based landing in both simulation and hardware (Section~\ref{sec:landing-perf}) using a 9g analog camera (selected for lower mass than a digital unit).

\subsection{Vision Controller: RGB Approach}
\label{sec:landing-vision-policy}
We used the ``RGB'' approach from the previous case study for vision-based control because, as indicated by Table~\ref{tab:leader-follower-analysis}, only the ``RGB'' controller comfortably meets the 20 Hz hardware control rates. 
Although ``RGB'' suffers from generalization issues, for landing, overfitting to the specific appearance is less problematic because the runway is usually static to the background. 
We therefore reuse the RGB imitation-learning setup from the leader-follower study (Section~\ref{sec:leader-follower}), but train in the simulation with the leader asset removed and the expert state-based controller's objective focusing on runway alignment and touchdown along the designated pad.

\subsection{Indoor Flying Arena Setup for Autonomous Landing}
\label{sec:experiment-setup}

All hardware trials were conducted in an indoor arena (40m$\times$20m$\times$5m) equipped with a blue landing pad (13m$\times$2m$\times$0.1m) placed at one end, as shown in Figure~\ref{fig:gsplat} Right. 
Each trial began with a human pilot manually piloting the aircraft to an initial position approximately 20\,m from the runway and 1.5\,m above ground, where the landing pad becomes roughly visible to the onboard analog camera. 
Upon reaching this position, control was switched to Autonomous Mode (Section~\ref{sec:hardware_system}), with the pilot instructed to immediately regain manual control if a flag was raised or any unsafe behavior was observed. 
%

\begin{figure}[htbp]
  \centering
  \includegraphics[width=\linewidth]{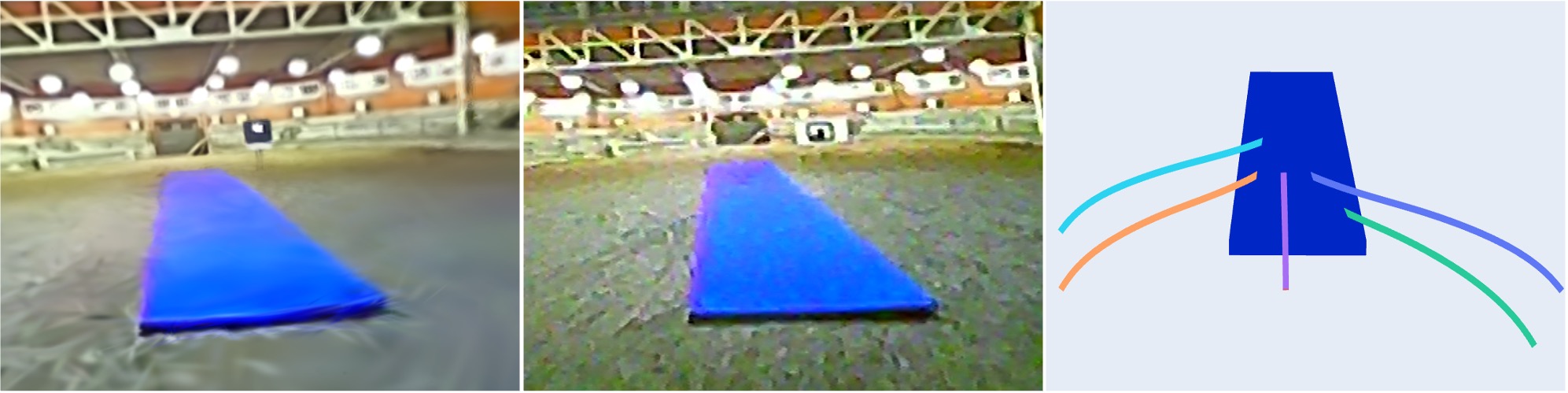}
  \caption{\small{Visualization for the autonomous landing case study: \textbf{Left} figure shows the GSplat-based simulation onboard view. \textbf{Middle} shows the real-world onboard view. \textbf{Right} shows a visualization of trajectories where our vision-based controller can successfully land from different initial positions.}}
  \label{fig:landing_traj}
\end{figure}

\subsection{Sim2Real Landing Performance}
\label{sec:landing-perf}

We performed 10 autonomous landing trials using our ``RGB'' vision-based controller in the real-world environment. 
Landing performance was evaluated based on two primary metrics: (1) \emph{landing success}, defined as touchdown within the bounds of the landing pad; and (2) \emph{Absolute Lateral Deviation (ALD)} from the runway centerline at touchdown. 
Given that our runway width is 2\,m, deviations less than or equal to 1\,m  is acceptable.
Because our flying arena currently lacks motion-capture, we assessed landing accuracy by coating the landing gear with powder and measuring the resulting touchdown marks on the runway.

For accurate simulation comparisons, we first recorded the image at the time of aircraft's initial Autonomous Mode engagement, then estimate the pose using our iGSplat model, and subsequently replayed each landing attempt in simulation with both our learned vision-based controller and the state-based expert controller, enabling direct comparison.

Results are summarized in Table~\ref{tab:controller}.  In simulation, both the state-based and vision-based controllers landed successfully in all ten cases, with mean ALD of 15cm and 37cm.  
Hardware trials achieved eight successful landings; the two failures (Runs 3 and 10) occurred after the aircraft was handed over with a steep nose-down attitude and the analog video suffered some flicker noise. 
The average ALD (41cm) of the real world trials are slightly larger than in simulation, this is likely due to difference between dynamics estimates and difference in image rendering quality.
Due to the lack of ground truth states, we could not run the state-based control in real world for comparison. 
However, additional simulation experiments show the vision-based controller can handle curved approaches and large initial offsets and different altitudes (Figure~\ref{fig:landing_traj}), but these were not flight-tested because of space constraints and bank-angle safety limits.

\begin{table}[htbp]
\centering
\caption{\small{
Controller Performance for Landing Case Study
}}
\label{tab:controller}
\scriptsize
\resizebox{\columnwidth}{!}{%
\begin{tabular}{|l|cc|cc|cc|}\toprule
      & \multicolumn{4}{c|}{Simulation} & \multicolumn{2}{c|}{Real World} \\ \hline 
      & \multicolumn{2}{c|}{State-based} & \multicolumn{2}{c|}{Vision-based} & \multicolumn{2}{c|}{Vision-based} \\ \hline 
    Run \# & Success? & ALD (cm) $\downarrow$ & Success? & ALD (cm) $\downarrow$ & Success? & ALD (cm) $\downarrow$ \\ \hline 
    1 & \cmark  & 42  & \cmark  & 54 & \cmark  & 35 \\
    2 & \cmark & 1 & \cmark & 4 & \cmark  & 45 \\
    3 & \cmark  & 14  & \cmark  & 38 & \xmark  & / \\
    4 & \cmark & 1 & \cmark & 12 & \cmark  & 40 \\
    5 & \cmark  & 27  & \cmark  & 32 & \cmark  & 40 \\
    6 & \cmark & 37 & \cmark & 74 & \cmark  & 70 \\
    7 & \cmark  & 13  & \cmark  & 21 & \cmark  & 20 \\
    8 & \cmark & 14 & \cmark & 79 & \cmark  & 80 \\
    9 & \cmark  & 2  & \cmark  & 32 & \cmark  & 78 \\
    10 & \cmark & 2 & \cmark & 23 & \xmark  & / \\

\hline

\end{tabular}
}
\end{table}

\section{Conclusion}  
\label{sec:conclusion}  

We introduced \emph{FalconWing}, an open-source platform for indoor fixed-wing UAV autonomy. 
FalconWing integrates a lightweight (150g) hardware stack with a software suite comprising photorealistic GSplat simulation and system-identified aircraft dynamics. 
We envision FalconWing to become the go-to aerial platform
for teaching and research.

We validate FalconWing on two challenging aerial case studies using only vision. 
In leader-follower visual tracking, de-coupled perception and control as well as domain-randomized GSplat training enable vision policies to generalize to unseen maneuvers and visual perturbations. 
In autonomous landing, an RGB-based controller trained purely in simulation transfers zero-shot to hardware, achieving an 80\% success rate over ten indoor trials without fine-tuning.
 
Future work includes: (i) identifying richer dynamics model that incorporates wind disturbances, flap/drag effects, and ground effect to narrow residual sim-to-real gaps; (ii) working on more challenging scenarios with controlled wind/lighting changes, temporary leader occlusions, rapid turning of the leader and sharper landing angles; (iii) distilling the most generalizable RGB+Mask controller into a lightweight model for hardware deployment. 

\section*{APPENDIX}

Standard lift equation is $L=\frac{1}{2}\rho V^2 SC \geq mg$ and the coordinated-turn relation is $R = \frac{V^2}{g tan(\phi)}$, assume $g{=}9.8$, our UMX wing area $S{=}0.076~\mathrm{m}^2$, air density $\rho{=}1.3~\mathrm{kg/m^3}$, UMX lift coefficient $C{=}0.6$, and bank angle $\frac{\pi}{6}$.





\bibliographystyle{IEEEtran}
\bibliography{reference}

\end{document}